\documentclass[11pt]{article}

\usepackage[preprint]{acl}

\usepackage{times}
\usepackage{latexsym}
\usepackage{amsfonts}
\usepackage{booktabs}
\usepackage{amssymb}
\usepackage[T1]{fontenc}

\usepackage[utf8]{inputenc}

\usepackage{microtype}

\usepackage{inconsolata}
\usepackage[table,xcdraw]{xcolor}
\usepackage{graphicx}
\usepackage{amsmath}
\usepackage{multirow}
\title{\textit{Every Token Leaves a Ripple in the Stream of Thought}: Eliciting Model-Internal Token Saliency for Chain-of-Thought Compression}

\author{
 \textbf{Tianyi Zhao\textsuperscript{1}},
 \textbf{Yinhan He\textsuperscript{1}},
 \textbf{Wendy Zheng\textsuperscript{1}},
 \textbf{Jundong Li\textsuperscript{1}},
 \textbf{Chen Chen\textsuperscript{1}}
\\
 \textsuperscript{1}University of Virginia
\\
\texttt{\{abs4dj,nee7ne,ncd9cf,jl6qk,zrh6du\}@virginia.edu}
}

\begin{document}
\maketitle
\begin{abstract}
Chain-of-thought (CoT) reasoning improves multi-step problem solving, but long reasoning traces inflate inference cost. Token-level CoT compression reduces this cost by pruning full reasoning chains into shorter traces for model adaptation, making token selection the central challenge. 
Existing methods often rely on external scorers or heuristic signals only indirectly tied to the model's internal answer computation.
We instead adopt a model-internal perspective: 
as the model forms an answer, each reasoning token induces a ripple in the residual stream whose effect on the answer reflects the token's contribution to the underlying computation.
Building on this view, we propose \textsc{MIST} (Model-Internal Saliency for Token-level CoT compression), which defines token importance along two
complementary axes: \emph{necessity}, the drop in answer
likelihood when a token's internal contribution is removed, and \emph{sufficiency}, the gain in answer likelihood
when that contribution alone is provided.
Combining the two yields a unified importance score for pruning.
Across four reasoning benchmarks and four models, \textsc{MIST} consistently outperforms baseline methods, suggesting that model-internal saliency provides an effective proxy for reasoning-token importance. 

\end{abstract}

\section{Introduction}

Chain-of-thought (CoT) reasoning~\citep{wei2022chain,kojima2022large,wang2023selfconsistency} improves the multi-step problem-solving ability of large language models (LLMs), but long reasoning traces increase latency, memory use, and serving cost. 
This has motivated work on CoT compression, which aims to preserve reasoning performance while shortening intermediate traces.
Previous methods obtain compact traces in several ways: prompting or distillation for concise
generation~\citep{xu2025cod}, adaptive reasoning
budgets~\citep{han2025token}, latent computation that replaces
explicit reasoning~\citep{shen2025codi}, and pruning tokens or
steps from a full trace~\citep{xia2025tokenskip,li2026making}.
We focus on the token pruning setting, where a full CoT trace is pruned to a subset of reasoning tokens that supervise the model to generate shorter chains. This setting makes the central bottleneck explicit: under a retention budget $\gamma$, which reasoning tokens should be kept?

\begin{figure}[t]
\centering
\includegraphics[width=\columnwidth]{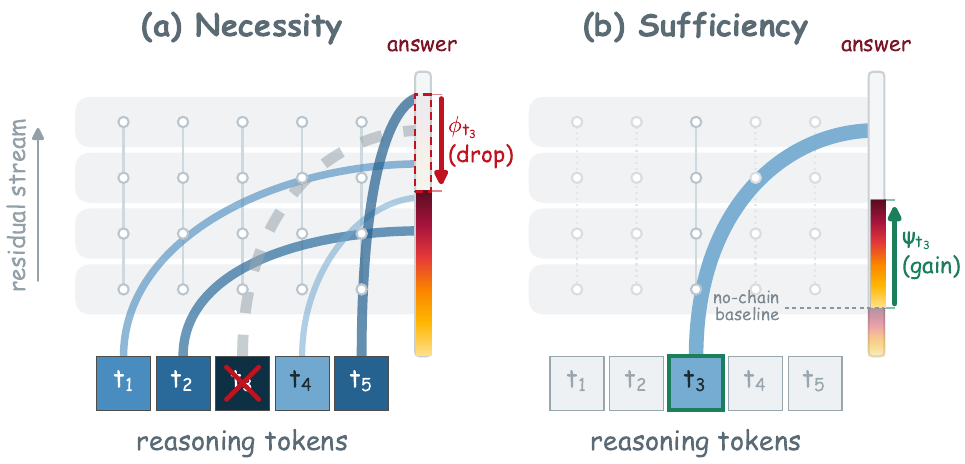}
\caption{\textbf{Conceptual overview of model-internal token saliency.}
For each reasoning token: \textbf{(a)~necessity} measures the decrease in answer likelihood when the token's residual contribution is removed; \textbf{(b)~sufficiency}
measures the increase in answer likelihood when that contribution is patched into a no-chain forward pass.
The two axes are combined to rank tokens, and the highest-scoring tokens are retained under compression.
}
\label{fig:teaser}
\end{figure}

Existing token-level compression methods answer this question using signals that are only indirectly tied to the target model's answer computation.
TokenSkip-style~\citep{xia2025tokenskip} methods, for example, rely on auxiliary scorers~\citep{pan2024llmlingua} whose rankings depend on the scorer's own training objective, supervision, and domain assumptions, rather than being derived from the target model's internal reasoning process.
While effective, they leave unexplored a more direct and principled path to measuring token importance: deriving it from what the target model itself relies on when forming the answer.
We therefore shift the focus from heuristic proxies to a complementary, model-internal perspective.
Under this view, a reasoning token is important when the target model’s answer computation depends on the information it carries in the residual stream, the internal substrate through which information propagates across layers. We view this residual stream as the model’s \emph{stream of thought}, in which each token leaves a ripple whose magnitude reflects its contribution to the answer.
Tokens that contribute little answer-relevant information are therefore natural candidates for removal under compression. 
This reframes token-level CoT compression as a model-internal saliency problem:
\emph{which reasoning tokens make the most internal contributions to the model's answer computation?}


We formalize this saliency along two complementary axes.
\textbf{(Q1) Necessity}: if a token's internal contribution is removed, how much does the ground-truth answer likelihood drop?
\textbf{(Q2) Sufficiency}: if only that token's internal information is provided, how much of the ground-truth answer likelihood is recovered?
Necessity reflects full-chain dependence: it captures tokens
whose erasure disrupts the chain's own computation.
Sufficiency reflects no-chain recoverability: it captures the contextual hinges whose residuals, given the query, recover answer-relevant information without the rest of the chain.
Each view is biased without the other: necessity admits low-information context that is not directly anchored to answer relevance under abundant retention budgets, while sufficiency alone underperforms under aggressive compression, when no single token's residual can substitute for the chain.
Combining the two yields a ranking that remains reliable across compression budgets. 

\begin{figure}[t]
\centering
  \includegraphics[width=\columnwidth]{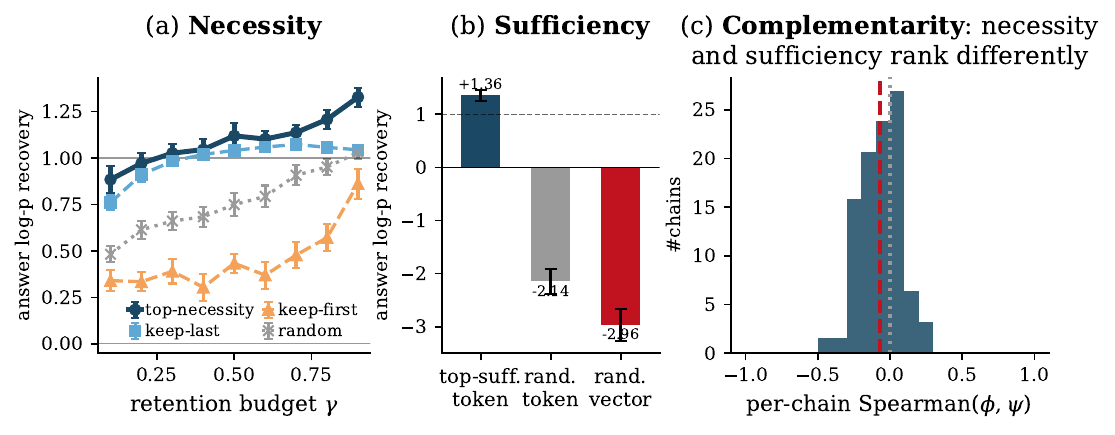}
  \caption{\textbf{Empirical characterization of dual-axis model-internal token saliency}.
  Results on 100 MATH reasoning chains with Qwen2.5-1.5B-Instruct.
  \textbf{(a)} Top-necessity retention consistently outperforms positional and random baselines across budgets. 
  \textbf{(b)} Top-sufficiency tokens recover substantially more answer signal than random and norm-matched controls. 
  \textbf{(c)} Necessity and sufficiency induce weakly correlated rankings (mean Spearman $\rho=-0.07$ and top-$30\%$ overlap $0.28$), indicating that they capture distinct yet complementary signals.
  }
  \label{fig:motivation}
\end{figure}

Building on this framing, we propose \textsc{MIST} (\underline{M}odel-\underline{I}nternal \underline{S}aliency for \underline{T}oken-level CoT compression).
For each reasoning token, \textsc{MIST} operationalizes the two axes as residual-stream interventions: 
necessity measures the effect of erasing a token's residual state from the full-chain computation, while sufficiency measures the answer signal recovered by patching that state into a no-chain forward pass.
Directly evaluating these interventions requires a separate pass for each token, scaling as $O(T)$ in chain length $T$ which makes token-wise scoring expensive.
We show that both axes admit a first-order Taylor
linearization in the residual stream, reducing each per-token
score to an inner product of gradients and activations, so a
single backward pass per axis yields all token scores.
\textsc{MIST} then combines the two scores into a unified token-importance score.
The highest-scoring tokens are retained to form compressed CoT traces, which are then used to adapt the model to generate shorter reasoning chains.

We evaluate \textsc{MIST} on four benchmarks spanning mathematical reasoning (GSM8K~\citep{cobbe2021training} and MATH~\citep{hendrycks2021measuring}) and general-domain reasoning (MMLU-Pro~\citep{wang2024mmlupro} and BIG-Bench Hard~\citep{suzgun2023challenging}), using Qwen2.5-1.5B-Instruct, Qwen2.5-7B-Instruct~\citep{qwen2.5}, Llama-3.1-8B-Instruct~\citep{grattafiori2024llama}, and Mistral-7B-Instruct-v0.3~\citep{Jiang2023Mistral7}.
Extensive results suggest that model-internal saliency serves as a reliable proxy for reasoning-token importance, and that the proposed method consistently improves CoT compression performance across settings.
In summary, our contributions are:
\begin{itemize}
    \item We formulate token-level CoT compression as a
model-internal saliency problem: rather than asking which tokens
appear important to heuristic proxies, we ask which reasoning
tokens contribute to the target model's own answer computation,
formalized along two complementary axes, necessity and sufficiency.
    \item We propose \textsc{MIST}, which operationalizes both
    axes as residual-stream interventions, derives first-order
    linearizations that reduce per-chain scoring to a single backward pass per axis, and combines the two axes into a unified token-importance score.
    \item Extensive experimental results across diverse datasets and models demonstrate the effectiveness of our proposed method.
\end{itemize}

\section{Related Work}

\paragraph{Efficient chain-of-thought reasoning.}
Chain-of-thought (CoT) prompting improves multi-step
reasoning~\citep{wei2022chain,kojima2022large,wang2023selfconsistency}
but inflates inference cost, motivating efficient-reasoning
methods spanning concise generation~\citep{xu2025cod,munkhbat2025self},
adaptive decoding budgets~\citep{han2025token}, length-controllable
fine-tuning~\citep{ma2025cot}, latent
reasoning~\citep{hao2024training,shen2025codi}, and token- or
step-level pruning of generated traces~\citep{xia2025tokenskip,li2026making}.
We focus on the pruning setting.

\paragraph{Token-level CoT pruning.}
Within the pruning setting,
TokenSkip~\citep{xia2025tokenskip} constructs compressed CoT supervision with an auxiliary LLMLingua-style token scorer~\citep{jiang2023llmlingua,pan2024llmlingua}, while step-level methods use signals such as entropy to skip generation~\citep{li2026making}.
More direct approaches include GoGI-Skip, which scores intermediate representations by gradient norm~\citep{zhuang2025accelerating}, and likelihood-preserving greedy deletion, which identifies removable tokens through repeated deletion evaluations~\citep{singh2026llms}.
In contrast, \textsc{MIST} derives a unified token-importance score from two complementary saliency measures in the model's residual stream, avoiding external scorers, single-axis heuristics, and costly iterative deletion.

\paragraph{Model-internal attribution and interventions.}
Mechanistic interpretability methods identify behaviorally relevant components such as neurons and attention heads through interventions such as causal mediation, causal tracing, and activation patching~\citep{vig2020investigating,meng2022locating,zhang2024towards,heimersheim2024use,syed2024attribution,ghandeharioun2024patchscopes}.
Residual-stream analysis and direct logit attribution decompose how internal components contribute to output logits~\citep{elhage2021mathematical,nostalgebraist2020logitlens}.
Inspired by this line of work, we define a token-saliency measure in the residual stream and use this model-internal signal to quantify the importance of reasoning tokens for CoT compression.

\section{\textsc{MIST}: Model-Internal Saliency for Token-level CoT Compression}
\label{sec:method}
We present \textsc{MIST}, a model-internal method for token-level CoT compression.
Given a query $x$, a reasoning chain $c=(t_1,\ldots,t_T)$ generated by the target model $M$, and the answer $a$, \textsc{MIST} assigns a saliency score to each chain token, retains the top-scoring tokens under a retention budget, and uses the resulting compressed chains to adapt the model for compact reasoning-trace generation.
The key idea is to first define what it means for a token to matter to the target model's own answer computation, and then quantify this importance from the model's internal states.

\begin{figure*}[t]
\centering
  \includegraphics[width=0.9\textwidth]{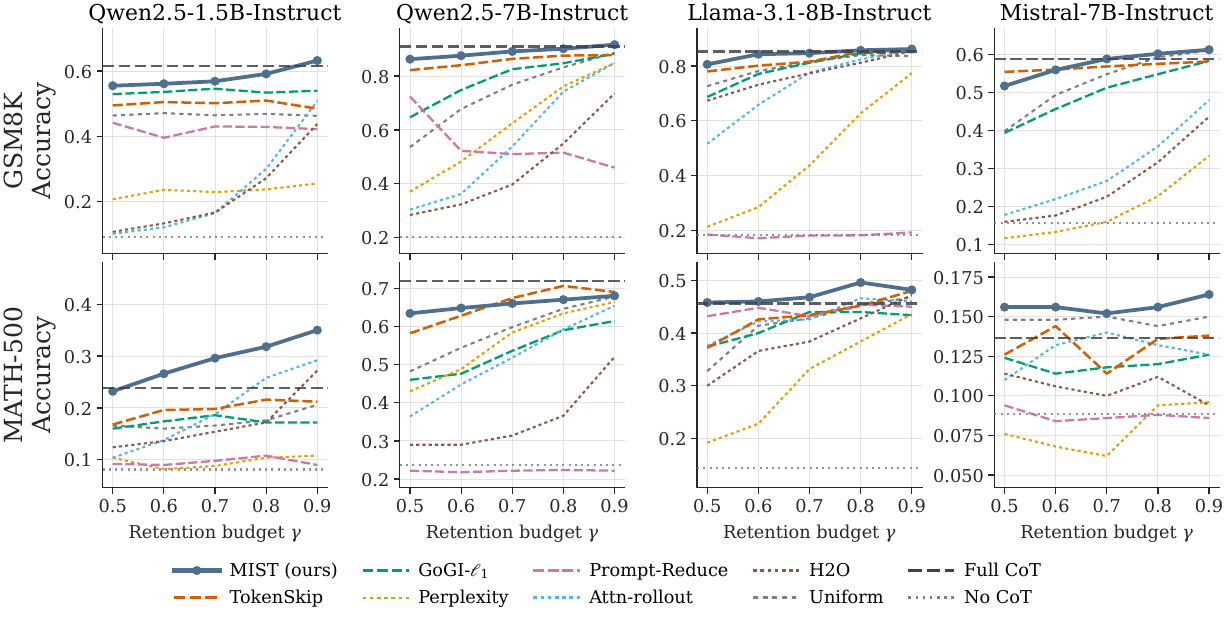}
  \caption{Accuracy across $\gamma \in \{0.5,...,0.9\}$ for
\textsc{MIST} and all baselines on GSM8K and MATH-500.}
  \label{fig:main_exp1}
\end{figure*}

\subsection{Problem Formulation: A Model-Internal Notion of Token Importance}
\label{sec:method:setup}
Token-level CoT compression requires a ranking of reasoning tokens. A natural objective is to keep the subset of tokens that best supports the model's answer computation:

\begin{equation}
    c_\gamma^\star = \arg\max_{c' \subseteq c,\ |c'|=\lceil \gamma T\rceil}\log p_M(a \mid x,c'),
\end{equation}
where \(\gamma\in(0,1]\) is the retention budget. Solving this objective directly is combinatorial, so practical compression methods reduce the problem to computing per-token importance and retaining the top-ranked subset.
In this work, we formulate this importance as \emph{model-internal saliency}: a token is salient if the target model's own answer computation depends on the internal information carried by that token.
To make this notion operational, we measure a token's internal contribution through its residual stream, a natural choice since the residual stream is the model's internal computational substrate.
We then define this saliency through two complementary axes.

\noindent\textbf{Necessity.}
Necessity quantifies how strongly the model's answer
computation is affected by removing a token's internal
contribution from the full chain.
Let $h_i$ denote the residual-stream states associated with token $i$. We operationalize this by zeroing $h_i$, and then quantify the resulting change in answer likelihood as the necessity saliency:
\begin{equation}
\resizebox{0.98\columnwidth}{!}{$
    {\phi}_i :=  \log p_M(a \mid x,c) - \log p_M\bigl(a \mid x,c^{h_i\to 0}\bigr).
$}
\end{equation}

\noindent\textbf{Sufficiency.}
Sufficiency measures how strongly an individual
token's internal state affects the answer
computation when introduced into a no-chain
forward pass.
We operationalize this by patching \(h_i\) into the no-chain forward pass at the final position and quantify the resulting change in answer likelihood as the sufficiency saliency:
\begin{equation}
\resizebox{0.98\columnwidth}{!}{$
    {\psi}_i := \log p_M\bigl(a \mid x,\mathrm{patch}(i)\bigr) - \log p_M(a \mid x,\varnothing).
$}
\end{equation}


\noindent\textbf{Why both axes? Informative and complementary views of token importance.}
The above definitions of necessity and sufficiency offer two behaviorally grounded notions of token importance. We further conduct an empirical evaluation on 100 gold MATH chains to show that both notions yield meaningful token-importance information in practice.
Retaining tokens ranked by exact necessity consistently outperforms positional and random baselines across compression budgets, indicating that necessity identifies tokens on which the full-chain answer computation depends (Fig.~\ref{fig:motivation}a). Conversely, patching the most sufficient tokens into no-chain computations recovers substantial answer-relevant information, whereas random-token and norm-matched controls do not (Fig.~\ref{fig:motivation}b). Despite their individual effectiveness, the two rankings are only weakly correlated and exhibit limited top-token overlap (Fig.~\ref{fig:motivation}c), suggesting that they capture largely distinct aspects of token importance.
Neither axis alone therefore provides a complete account of token importance, motivating their combination in our dual-axis saliency measure.
%

\subsection{Quantifying Model-Internal Saliency in Practice}
\label{sec:method:taylor}
The formulations above provide conceptually clean measures of token saliency, but their naive computation is prohibitively expensive. 
Evaluating each token would require an additional forward pass to instantiate either \(c^{h_i\to 0}\) or \(\mathrm{patch}(i)\), yielding an \(O(T)\) cost per chain
that is impractical at scale.
Moreover, simultaneously intervening a token's residual
states across all layers may cause compounded
perturbations that substantially disrupt the original
computation.
We therefore consider interventions at individual layers and derive efficient layerwise proxies by linearizing $\log p_M$ around the unperturbed forward passes.
This yields a practical first-order scorer computable with two backward passes per chain, independent of $T$.

Let $h_i^{(l)}$ for $l \in \{1, \ldots, L\}$ denote token $i$'s residual at layer $l$, with $h_i^{(l),\text{src}}$ and $h_{\text{final}}^{(l),\text{tgt}}$ denoting the unperturbed source and target activations at layer $l$.
The necessity intervention $c^{h_i\to 0}$ perturbs $h_i^{(l), \mathrm{src}}$ along the direction $-h_i^{(l), \mathrm{src}}$;
the sufficiency intervention $\mathrm{patch}(i)$ perturbs $h_{\mathrm{final}}^{(l), \mathrm{tgt}}$ along the direction $h_i^{(l), \mathrm{src}} -h_{\mathrm{final}}^{(l), \mathrm{tgt}}$.
The answer log-likelihood $\log p_M$
is differentiable in the residual activations, so both saliencies admit a first-order Taylor expansion along these directions.

\noindent\textbf{Necessity term.}
The first-order expansion of $\log p_M^{\mathrm{src}}$ along $-h_i^{(l),\mathrm{src}}$ identifies the per-layer main term:
\begin{equation}
    \widehat\phi_i^{(l)} \;:=\; \bigl\langle \nabla_{h_i^{(l)}}\log p_M^{\mathrm{src}},\;
        h_i^{(l),\mathrm{src}} \bigr\rangle,
    \label{eq:phi-layer}
\end{equation}
Both the gradient and the activation in $\widehat\phi_i^{(l)}$ are read from the same source forward pass: one backward through $\log p_M^{\mathrm{src}}$ yields $\nabla_{h_i^{(l)}}\log p_M^{\mathrm{src}}$ at every $(i, l)$ pair simultaneously, so the per-chain cost on the necessity axis is one forward and one backward.

\noindent\textbf{Sufficiency term.}
The first-order expansion of $\log p_M^{\mathrm{tgt}}$ along $h_i^{(l),\mathrm{src}} - h_{\mathrm{final}}^{(l),\mathrm{tgt}}$ identifies the per-layer main term:
\begin{equation}
    \widehat\psi_i^{(l)} \;:=\; \bigl\langle
        \nabla_{h_{\mathrm{final}}^{(l)}}\log p_M^{\mathrm{tgt}},\;
        h_i^{(l),\mathrm{src}} - h_{\mathrm{final}}^{(l),\mathrm{tgt}} \bigr\rangle,
    \label{eq:psi-layer}
\end{equation}
The gradient is taken at the target's final position; one backward through $\log p_M^{\mathrm{tgt}}$ yields these gradients across all layers $l$, and the source activations $h_i^{(l), \mathrm{src}}$ are reused from the necessity pass.

The full Taylor expansion and remainder bounds for both terms are given in Appendix~\ref{app:proofs}.

\subsection{The Unified \textsc{MIST} Score}
\label{sec:method:score}
The per-layer terms $\widehat\phi_i^{(l)}$ and $\widehat\psi_i^{(l)}$ are building blocks;
the \textsc{MIST} score combines them into a unified per-token saliency signal by first aggregating across layers using answer-aware layer weights along each axis, and then combining the two axes.

\noindent\textbf{Per-axis aggregation.}
Different layers contribute differently to the answer computation; we therefore adopt a logit-lens-induced~\citep{nostalgebraist2020logitlens} layer weighting:
each transformer layer $l$ writes an additive update
$h^{(l)}_t - h^{(l-1)}_t$ into the residual-stream state at
position $t$, and the inner product of this update with the
unembedding row $W_U[a]$ corresponding to the gold answer token,
\begin{equation}
    \bar w_l \;=\; \frac{1}{T} \sum_{t=1}^{T}
        \bigl\langle h^{(l)}_t - h^{(l-1)}_t,\; W_U[a] \bigr\rangle,
    \label{eq:logit-lens-weight}
\end{equation}
captures how much that layer's update pushes the chain-averaged hidden state toward the answer; we retain only positive alignment by setting $\bar c_l=\max(\bar w_l,0)$.
Aggregating each axis with this shared weight yields the per-token saliencies along the two axes:
\begin{equation}
    \widehat\phi_i \;=\; \sum_l \bar c_l \cdot
        \widehat\phi_i^{(l)},
    \quad
    \widehat\psi_i \;=\; \sum_l \bar c_l \cdot \widehat\psi_i^{(l)}.
    \label{eq:axis-agg}
\end{equation}

\noindent\textbf{Combining the two axes.}
Finally, we combine the two axes as\footnote{$\widehat\phi_i$ and $\widehat\psi_i$ are normalized before combining; see App.~\ref{app:proofs:normalize} for details.}:
\begin{equation}
    S^{\textsc{MIST}}_i \;=\;
        \alpha \cdot \widehat\phi_i + (1-\alpha) \cdot \widehat\psi_i,\quad \alpha \in [0, 1].
    \label{eq:mist-score}
\end{equation}
where $\alpha$ is a hyperparameter. Tokens ranked by $S^{\textsc{MIST}}_i$ form the compressed chains that supervise the fine-tuning of $M$ at retention budget $\gamma$.

\section{Experiments}

\begin{table}[t]
\centering
\renewcommand{\arraystretch}{1.08}
\resizebox{\columnwidth}{!}{%
\begin{tabular}{c c l c c}
\toprule
\textbf{Dataset} & \textbf{Model} & \textbf{Method} 
& \textbf{$\Delta$ Acc. (\%)} & \textbf{Comp. rate} \\
\midrule

\multirow{12}{*}{{GSM8K}}
& \multirow{6}{*}{\rotatebox{90}{Qwen2.5-7B-Inst.}}
& $\star$ MIST   & $\downarrow 2.4$  & $21.3\%$ \\
& & tokenskip     & $\downarrow 4.3$  & $20.0\%$ \\
& & gogi\_l1      & $\downarrow 11.9$ & $8.9\%$  \\
& & perplexity    & $\downarrow 29.2$ & $17.6\%$ \\
& & attn\_rollout & $\downarrow 35.1$ & $21.4\%$ \\
& & h2o           & $\downarrow 45.2$ & $22.0\%$ \\

\cmidrule(lr){2-5}

& \multirow{6}{*}{\rotatebox{90}{Mistral-7B-Inst.}}
& $\star$ MIST   & $\downarrow 1.3$  & $13.4\%$ \\
& & tokenskip     & $\downarrow 2.1$  & $10.4\%$ \\
& & gogi\_l1      & $\downarrow 9.2$ & $14.8\%$ \\
& & perplexity    & $\downarrow 39.5$ & $25.6\%$ \\
& & attn\_rollout & $\downarrow 28.8$ & $22.5\%$ \\
& & h2o           & $\downarrow 32.6$ & $23.0\%$ \\

\midrule

\multirow{12}{*}{MATH-500}
& \multirow{6}{*}{\rotatebox{90}{Qwen2.5-1.5B-Inst.}}
& $\star$ MIST   & $\uparrow 5.3$  & $9.3\%$  \\
& & tokenskip     & $\downarrow 4.1$ & $10.2\%$ \\
& & gogi\_l1      & $\downarrow 6.6$ & $3.9\%$  \\
& & perplexity    & $\downarrow 14.1$ & $8.5\%$  \\
& & attn\_rollout & $\downarrow 4.2$ & $0.0\%$  \\
& & h2o           & $\downarrow 6.7$ & $4.7\%$  \\

\cmidrule(lr){2-5}

& \multirow{6}{*}{\rotatebox{90}{Llama-3.1-8B-Inst.}}
& $\star$ MIST   & $\uparrow 1.2$   & $10.6\%$ \\
& & tokenskip     & $\downarrow 2.3$  & $13.8\%$ \\
& & gogi\_l1      & $\downarrow 3.8$  & $7.2\%$  \\
& & perplexity    & $\downarrow 14.1$ & $11.1\%$ \\
& & attn\_rollout & $\downarrow 2.6$  & $8.0\%$  \\
& & h2o           & $\downarrow 6.6$ & $10.0\%$ \\

\bottomrule
\end{tabular}%
}
\caption{
Performance and compression comparison on GSM8K and MATH-500.
Each entry is the mean over $\gamma \in \{0.5, 0.6, 0.7, 0.8, 0.9\}$ relative to the full-chain baseline. 
$\Delta$Acc.\,(\%): relative accuracy change ($\downarrow$ drop, $\uparrow$ gain); Comp.\,rate: fraction by which
the LoRA-adapted model's {inference-time generated
tokens} are reduced relative to the same full-chain
baseline.
}
\label{tab:main-compression-results}
\end{table}

\subsection{Experimental Setup}
\label{sec:exp:setup}

\paragraph{Datasets and Models.}
We evaluate on four reasoning benchmarks spanning mathematical
reasoning (GSM8K~\citep{cobbe2021training}; MATH~\citep{hendrycks2021measuring}) and general-domain reasoning
(MMLU-Pro~\citep{wang2024mmlupro}; BIG-Bench Hard~\citep{suzgun2023challenging}).
Per-dataset split sizes, preprocessing, and self-generation
hyperparameters are deferred to Appendix~\ref{app:datasets}.
%
We evaluate four models covering
three families and three scales:
Qwen2.5-1.5B-Instruct and Qwen2.5-7B-Instruct~\citep{qwen2.5},
Llama-3.1-8B-Instruct~\citep{grattafiori2024llama}, and
Mistral-7B-Instruct-v0.3~\citep{Jiang2023Mistral7}.

\paragraph{Baselines.}
We compare \textsc{MIST} against nine baselines across three categories.
\underline{Training-free baselines}:
{prompt-reduce}~\citep{xia2025tokenskip} prompts the frozen
target model to compress its own chain without finetuning.
\underline{Naive baselines}:
{full-chain}, {no-chain}, and {uniform}
share the pipeline but use trivial token selection: no compression,
chain removal, and uniform retention, respectively.
\underline{Token scorer baselines}:
{TokenSkip}~\citep{xia2025tokenskip} scores tokens with the
external LLMLingua-2 classifier,
{GoGI}~\citep{zhuang2025accelerating} with the single-layer
gradient norm,
{perplexity} with per-token self-information~\citep{jiang2023llmlingua},
and {attention rollout}~\citep{abnar2020quantifying} and
{H2O}~\citep{zhang2023ho} with two attention-based proxies.
Per-baseline details are in
Appendix~\ref{app:baselines}.

\begin{figure}[t]
\centering
  \includegraphics[width=0.9\columnwidth]{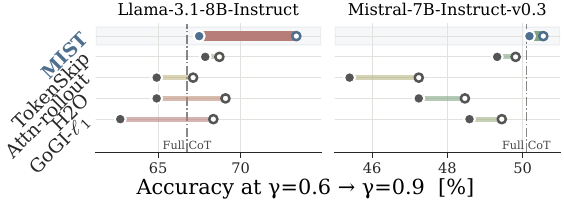}
  \caption{Results on BIG-Bench Hard.}
  \label{fig:bbh}
\end{figure}

\begin{figure}[t]
\centering
  \includegraphics[width=0.9\columnwidth]{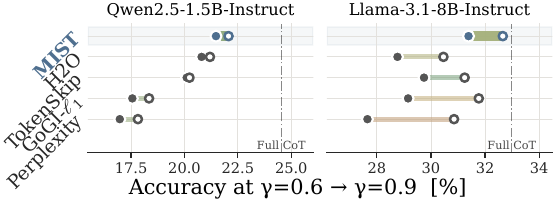}
  \caption{Results on MMLU-Pro.}
  \label{fig:mmlu}
\end{figure}

\paragraph{Training and evaluation protocol.}
We adopt the standard procedure from
TokenSkip~\citep{xia2025tokenskip}.
Given a target model $M$ and a training set, we (i) sample self-generate reasoning chains from $M$ and keep those that yield the correct answer, (ii) for each retention budget $\gamma\in\{0.5,0.6,\ldots,1.0\}$ in a grid, score each chain with \textsc{MIST} and form a compressed chain by retaining the top-$\lceil \gamma T \rceil$ tokens, and (iii) fine-tune $M$ with a LoRA adapter (rank~8, $\alpha=16$, learning rate $5\!\times\!10^{-5}$, three epochs, effective batch size~8) on the mixture of compressed chains across all $\gamma$ values, so that the same adapter can be decoded at any retention budget at inference.
It should be noted that since $\gamma$ controls the supervision tokens retained during SFT rather than imposing a hard length constraint at inference, the adapter is decoded freely and the actual inference-time compression rate ({Comp.\,rate} in
Table~\ref{tab:main-compression-results}) is typically smaller
than $1-\gamma$.
Additional details are provided in Appendix~\ref{app:protocol}.

\subsection{Main Results}
\label{sec:exp:main-results}
We report main results in Figure~\ref{fig:main_exp1} and
Table~\ref{tab:main-compression-results}.

\paragraph{Model-internal saliency outperforms external scorers.}
{TokenSkip} is the strongest baseline overall, but it relies on an
auxiliary scorer whose importance signal is not derived from the target model's own answer computation. Across the evaluated datasets and models, \textsc{MIST} achieves better performance than
{TokenSkip}. Averaged over \(\gamma\), \textsc{MIST} incurs only a
1.3-2.4 percentage-point (pp) accuracy drop on GSM8K with Qwen2.5-7B
and Mistral-7B, compared with 2.1-4.3 pp for {TokenSkip}. The
gap is even larger on MATH-500: \textsc{MIST} improves average accuracy by 5.3 pp on Qwen2.5-1.5B and 1.2 pp on Llama-3.1-8B, whereas {TokenSkip} drops by 4.1 pp and 2.3 pp, respectively. This suggests that the target model's own internal computations provide a reliable signal for identifying the reasoning tokens it actually relies on.

\paragraph{A principled internal scorer outperforms internal heuristics.}
The remaining baselines use target-model signals such as gradients, perplexity, or attention weights, but map these signals to token importance through heuristic scoring rules.
For instance, GoGI is the closest internal-gradient baseline, but it treats gradient norm at a preselected layer as a heuristic proxy for token importance, without deriving it from any explicit definition of saliency.
\textsc{MIST}, by contrast, operationalizes token importance through necessity and sufficiency interventions and derives its scorer from their first-order effects on the answer likelihood.
The gradient appears in \textsc{MIST} only multiplied by
the activation and aggregated across all layers; the bare gradient norm used by {GoGI} can be viewed as a strict simplification of this construction along the necessity axis. 
The empirical gaps further highlight the advantage of \textsc{MIST}'s principled saliency construction.
On GSM8K, {GoGI} incurs 11.9 and 9.2 pp accuracy drops with Qwen2.5-7B and Mistral-7B, respectively, compared with only 2.4 and 1.3 pp for \textsc{MIST}. 
On MATH-500, {GoGI} drops by 6.6 pp with Qwen2.5-1.5B and 3.8 pp
with Llama-3.1-8B, while \textsc{MIST} improves over full CoT by
5.3 and 1.2 pp. 
Other internal or model-derived heuristics degrade further, with perplexity, attention rollout, and {H2O} reaching drops as large as 39.5, 28.8, and 32.6 pp, respectively
(Table~\ref{tab:main-compression-results}).


\paragraph{\textsc{MIST} achieves favorable accuracy-compression trade-off.}
\textsc{MIST} achieves competitive compression while incurring the smallest reasoning performance loss.
On GSM8K with Qwen2.5-7B, \textsc{MIST} achieves a 21.3\% reduction in chain length, compared with 20.0\% for {TokenSkip}, while incurring only a 2.4-pp accuracy drop compared with 4.3 pp for {TokenSkip}.
Methods with internal heuristics signals, such as {perplexity}, {attention rollout}, and {H2O}, can achieve similar or higher compression rates but suffer much larger accuracy drops.
This trade-off suggests that, under the same retention budget,
the tokens selected by \textsc{MIST} carry a greater share of
the answer-relevant signal, allowing the adapter to recover
chain-level accuracy from a shorter trace.

\begin{figure}[t]
\centering
  \includegraphics[width=0.65\columnwidth]{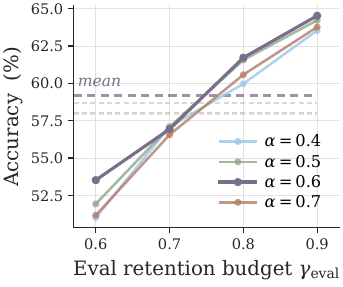}
  \caption{Sensitivity analysis for $\alpha$ on GSM8K with Qwen2.5-1.5B-Instruct.}
  \label{fig:alpha_ablation}
\end{figure}

\begin{figure}[h]
\centering
  \includegraphics[width=0.7\columnwidth]{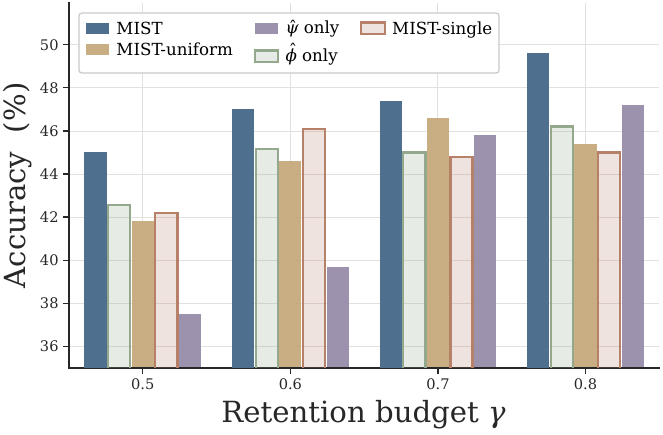}
  \caption{Component ablation of \textsc{MIST} on MATH-500
with Llama-3.1-8B-Instruct.}
  \label{fig:component_ablation}
\end{figure}

\begin{figure*}[t]
  \includegraphics[width=0.9\textwidth]{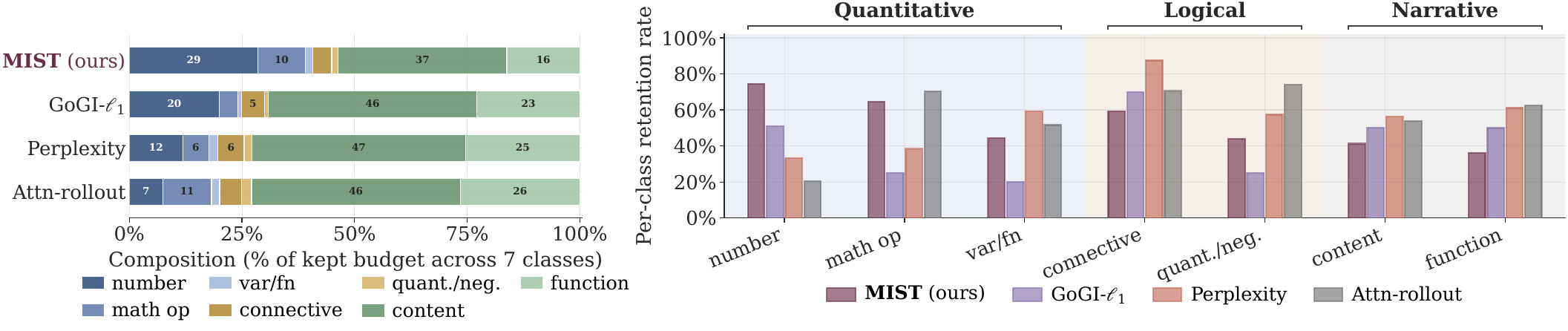}
  \caption{
  \textbf{What each scorer keeps under compression
(GSM8K, Qwen2.5-1.5B-Instruct).} Tokens classified using the
taxonomy in Table~\ref{tab:tax}.
  \textbf{Left.} How each method allocates its 50\% retention budget across the 7 token types.
  \textbf{Right.} Per-class retention rate: for each class (x-axis), the fraction of class tokens retained by each method.
  }
  \label{fig:token}
\end{figure*}

\subsection{Generalization Beyond Mathematical Reasoning}
\label{sec:exp:generalization}
The main results in \S\ref{sec:exp:main-results} establish
\textsc{MIST}'s advantage on two mathematical-reasoning benchmarks, GSM8K and MATH-500. We next ask whether this advantage extends beyond mathematical reasoning.
To this end, we further evaluate on two general-domain benchmarks: MMLU-Pro and BBH.

As shown in Figures~\ref{fig:bbh} and~\ref{fig:mmlu}, \textsc{MIST} remains the strongest method on both benchmarks and across the evaluated models.
On MMLU-Pro, \textsc{MIST} consistently outperforms external and
heuristic baselines, indicating that model-internal saliency remains effective in multi-domain, knowledge-intensive settings. On BBH, \textsc{MIST} also stays closest to, and in some settings surpasses, the {full CoT} reference,
while heuristic baselines fall noticeably behind. These results suggest that \textsc{MIST} is not merely exploiting math-specific token patterns; it effectively captures answer-relevant internal contributions across broader domains and more heterogeneous reasoning formats.
Additional results are reported in App.~\ref{app:additional-results}.

\subsection{Ablation Study}
\label{sec:exp:ablation}

\paragraph{Sensitivity analysis.}
We examine the sensitivity of \textsc{MIST} to the mixing coefficient $\alpha$ in Eq.~\eqref{eq:mist-score}, using Qwen2.5-1.5B-Instruct on GSM8K. 
Figure~\ref{fig:alpha_ablation} compares \(\alpha\in\{0.4,0.5,0.6,0.7\}\) across retention budgets.
Performance is stable across this range, indicating that the blend is not knife-edge sensitive to $\alpha$. We fix $\alpha=0.6$ in all main experiments as it achieves the best overall performance

\paragraph{Each component contributes to the unified \textsc{MIST} quality.}
We ablate the three design choices that distinguish \textsc{MIST} from one-axis or single-layer simplifications: (i)~using both saliency axes, by removing one at a time ($\widehat\varphi$-only, $\widehat\psi$-only); (ii)~the layer weighting in Eq.~\ref{eq:axis-agg}, by replacing $\bar c_l$ with a uniform per-layer weight (\textsc{MIST}-uniform); and (iii)~the multi-layer aggregation, by computing each axis from only the penultimate layer (\textsc{MIST}-single).
Figure~\ref{fig:component_ablation} reports accuracy on MATH-500 with Llama-3.1-8B-Instruct across $\gamma\in\{0.5, 0.6, 0.7, 0.8\}$. Full \textsc{MIST} achieves the best performance at every retention budget.
The single-axis ablations lose the most:
$\widehat\psi$-only drops by up to $7.5$~pp at the most aggressive budget ($\gamma{=}0.5$: $37.5\%$ vs.\ \textsc{MIST}'s $45.0\%$), and $\widehat\varphi$-only loses up to $3.4$~pp at $\gamma{=}0.8$ ($46.2\%$ vs.\ $49.6\%$).
\textsc{MIST}-uniform and \textsc{MIST}-single each lose between $0.8$ and $4.6$~pp depending on $\gamma$, confirming that neither the weighting nor the multi-layer aggregation is interchangeable with a uniform-weight or single-layer simplification. The three design choices combine to yield \textsc{MIST}'s final quality.

\subsection{Token Type Analysis}
\label{sec:exp:token-type}

To understand what each method actually keeps, we partition
every token into one of seven role classes
grouped into three macro-categories (Table~\ref{tab:tax}):
Quantitative (number, math operator, variable / function), Logical (connective, quantifier / negation), and Narrative (content, function). We then measure each method's
budget composition (left) and per-class retention rate (right) on Qwen2.5-1.5B-Instruct + GSM8K at $\gamma=0.5$
(Figure~\ref{fig:token}).


\begin{table}[h]
  \centering
  \setlength{\tabcolsep}{3.5pt}
  \renewcommand{\arraystretch}{1.15}
  \definecolor{taxquant}{HTML}{3B5984}
  \definecolor{taxlogic}{HTML}{B89243}
  \definecolor{taxnarr}{HTML}{6F9876}
  \small
  \resizebox{0.7\columnwidth}{!}{%
  \begin{tabular}{@{}l p{2.95cm} p{2.35cm}@{}}
    \toprule
    \textbf{Type} & \textbf{Definition} & \textbf{Examples} \\
    \midrule
    \multicolumn{3}{@{}l@{}}{\textcolor{taxquant}{\textbf{Quantitative}}} \\
    \midrule
    \textbf{number}      & integers, decimals, fractions, percentages     & \texttt{15}, \texttt{3.14} \\
    \textbf{math op}     & arithmetic operators and relations             & $+$, $-$, $\times$, $\div$, $=$, $\geq$ \\
    \textbf{var/fn}      & single-letter variables and math functions     & \texttt{x}, \texttt{y}, \texttt{sin}, \texttt{log} \\
    \midrule
    \multicolumn{3}{@{}l@{}}{\textcolor{taxlogic}{\textbf{Logical}}} \\
    \midrule
    \textbf{connective}  & causal / inferential / sequential glue         & \texttt{so}, \texttt{if}, \texttt{then}, \texttt{because} \\
    \textbf{quant./neg.} & quantifiers and negation markers               & \texttt{all}, \texttt{some}, \texttt{no}, \texttt{not} \\
    \midrule
    \multicolumn{3}{@{}l@{}}{\textcolor{taxnarr}{\textbf{Narrative}}} \\
    \midrule
    \textbf{content}     & nouns, verbs, adjectives, adverbs              & \texttt{Janet}, \texttt{eggs}, \texttt{calculate} \\
    \textbf{function}    & articles, prepositions, pronouns, auxiliaries  & \texttt{the}, \texttt{of}, \texttt{is}, \texttt{she} \\
    \bottomrule
  \end{tabular}
  }
  \caption{Token taxonomy used in the analysis.}
  \label{tab:tax}
\end{table}

Figure~\ref{fig:token} reports two complementary views. The left panel shows how each method allocates its retained-token budget across token classes. MIST allocates a larger share of its budget to quantitative tokens than heuristic baselines do: \(29\%\) to numbers and \(10\%\) to mathematical operators, while still retaining content and function words for local coherence. 
Within these classes, \textsc{MIST} also retains
the highest fraction: $75\%$ of all numbers and $66\%$ of all
mathematical operators (Figure~\ref{fig:token} right). In contrast, {attention rollout} and {perplexity} allocate only $7\%$ and $12\%$ of their budgets to numbers, retaining $22\%$ and $33\%$ of number tokens, respectively.

\textsc{MIST} is therefore the only scorer that both over-allocates budget to Quantitative tokens and achieves the highest within-class retention for them. Attention rollout concentrates on frequent operators and function words
that often act as attention hubs, while perplexity tends to conflate surprisal, retaining high-surprisal but information-light tokens such as \emph{therefore} and \emph{so}.
This pattern is consistent with the diagnosis in \S\ref{sec:exp:main-results}: heuristic scorers concentrate on syntactically prominent but information-light tokens, while \textsc{MIST}'s saliency-derived ranking aligns with the tokens that carry the answer-relevant computation.

\begin{figure}[h]
  \includegraphics[width=0.98\columnwidth]{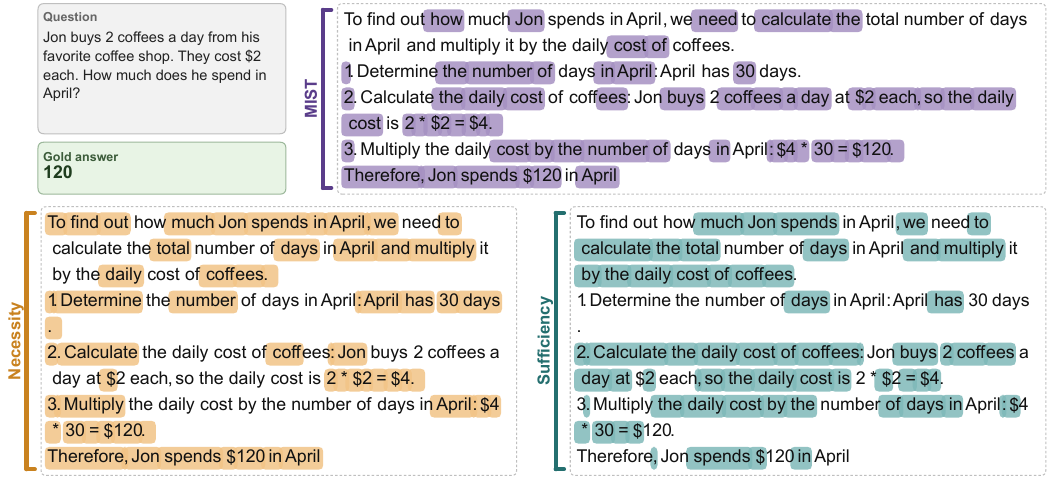}
  \caption{Per-token MIST saliency on a GSM8K chain}
  \label{fig:case}
\end{figure}

\subsection{Case Study}

Figure~\ref{fig:case} visualizes \textsc{MIST}'s saliency on a three-step GSM8K chain.
Necessity $\widehat\phi$ assigns high saliency to tokens that the full-chain computation depends on, concentrating on the
intermediate numerical facts ($\$4$, $30$) and the final answer $\$120$.
Sufficiency $\widehat\psi$ instead highlights a related
but distinct set of tokens, including those associated with buying, spending, currency, and equality, whose residual states carry answer-relevant information when supplied to the no-chain computation.
The two views overlap on some answer-bearing tokens
but also emphasize different parts of the reasoning trace.
Combining these signals, the unified \textsc{MIST} score retains tokens around the arithmetic computation $\$4 \times 30 = \$120$ together with the answer-bearing phrase ``spends $\$120$'', preserving both quantitative anchors and the local semantic context that binds them, and matching the class-level allocation reported in \S\ref{sec:exp:token-type}.

\section{Conclusion}
We present \textsc{MIST}, a model-internal method for token-level CoT compression. \textsc{MIST} defines token
importance operationally through two complementary interventions instantiated on the residual stream, necessity and sufficiency, and converts both into a tractable per-token scorer via first-order Taylor approximations, reducing the cost of token-wise interventions to two backward passes per chain. 
Extensive empirical evaluations across diverse datasets and models demonstrate the effectiveness of our proposed method.

\section*{Limitations}
\paragraph{Model scale and gradient access.}
Our experiments use 1.5B to 8B instruction-tuned models. Whether the saliency signal remains similarly concentrated in larger models is not established.
Besides, \textsc{MIST} requires access to the target model's internal activations and gradients, so it does not apply to black-box API-only systems.

\paragraph{Evaluation domain.}
Our four benchmarks span mathematical and general-domain reasoning. Future work can extend the framework to open-ended
generation, code generation, or multi-turn dialogue to further validate the effectiveness of the proposed method.


\paragraph{Approximate intervention scores.}
\textsc{MIST} approximates the necessity and sufficiency interventions with their first-order Taylor terms in the residual stream.
Because transformer computations are nonlinear, higher-order
interactions among tokens, layers, and heads are not captured, so
the \textsc{MIST} score is a model-internal saliency estimator
rather than an exact causal decomposition.

\bibliography{custom}

\appendix

\section{Dataset Details}
\label{app:datasets}

Table~\ref{tab:dataset-splits} summarizes the per-dataset split sizes
used throughout our evaluation pipeline. Below we describe per-dataset
preprocessing and self-generation hyperparameters.

\begin{table}[h]
\centering
\resizebox{0.5\columnwidth}{!}{%
\begin{tabular}{lrr}
\toprule
\textbf{Dataset} & \textbf{Train} & \textbf{Test} \\
\midrule
GSM8K     & 7{,}473  & 1{,}319  \\
MATH      & 7{,}500  & 500    \\
MMLU-Pro  & 10{,}501 & 1{,}531 \\
BBH-MC    & 3{,}258  & 816  \\
\bottomrule
\end{tabular}%
}
\caption{Dataset split sizes used in our evaluation pipeline.}
\label{tab:dataset-splits}
\end{table}

\paragraph{GSM8K.}
We use the standard \texttt{train} and \texttt{test} splits from the
GSM8K release without further filtering.

\paragraph{MATH.}
The training set is the concatenation of the seven
Hendrycks~\citep{hendrycks2021measuring} subjects;
the test set is the canonical MATH-500 subset.

\paragraph{MMLU-Pro.}
The original release does not provide
a standard training split, supplying only a 12{,}032-example test
split and a 70-example few-shot validation set. We therefore
randomly partition the 12{,}032 examples into 10{,}501 training
and 1{,}531 evaluation examples.

\paragraph{BBH-MC.}
We use the 17 letter-multiple-choice subtasks of BIG-Bench
Hard~\citep{suzgun2023challenging}, excluding the 10 binary or
free-form subtasks. BBH does not provide a standard training
split, supplying only a test set per subtask, so we randomly
partition the available examples into 3{,}258 training and 816
evaluation examples (80/20 split).

\begin{table}[h]
\centering
\resizebox{\columnwidth}{!}{%
\begin{tabular}{llr}
\toprule
\textbf{Subtask} & \textbf{Letters} & \textbf{N} \\
\midrule
\texttt{date\_understanding}                       & A--F & 250 \\
\texttt{disambiguation\_qa}                        & A--C & 250 \\
\texttt{geometric\_shapes}                         & A--K & 250 \\
\texttt{hyperbaton}                                & A--B & 250 \\
\texttt{logical\_deduction\_three\_objects}        & A--C & 250 \\
\texttt{logical\_deduction\_five\_objects}         & A--E & 250 \\
\texttt{logical\_deduction\_seven\_objects}        & A--G & 250 \\
\texttt{movie\_recommendation}                     & A--D & 250 \\
\texttt{penguins\_in\_a\_table}                    & A--E & 146 \\
\texttt{reasoning\_about\_colored\_objects}        & A--R & 250 \\
\texttt{ruin\_names}                               & A--D & 250 \\
\texttt{salient\_translation\_error\_detection}    & A--F & 250 \\
\texttt{snarks}                                    & A--B & 178 \\
\texttt{temporal\_sequences}                       & A--D & 250 \\
\texttt{tracking\_shuffled\_objects\_three\_objects} & A--C & 250 \\
\texttt{tracking\_shuffled\_objects\_five\_objects}  & A--E & 250 \\
\texttt{tracking\_shuffled\_objects\_seven\_objects} & A--G & 250 \\
\midrule
\multicolumn{2}{l}{\textbf{Total}}                 & \textbf{4{,}074} \\
\bottomrule
\end{tabular}%
}
\caption{The 17 subtasks of BIG-Bench
Hard used in our \textsc{MIST} evaluation. ``Letters'' shows the
answer-letter range; ``N'' is the original number of examples
per subtask before our 80/20 partition.}
\label{tab:bbh-subtasks}
\end{table}

\section{Baseline Details}
\label{app:baselines}
This appendix details the nine baselines compared in
Section~\ref{sec:exp:setup}.

\paragraph{\texttt{Prompt-reduce}.}
The frozen target model is prompted with the chain-of-thought
instruction together with the reduction directive from TokenSkip
\citep[Table~1]{xia2025tokenskip}, which appends
\texttt{please reduce $(1{-}\gamma)\!\times\!100\%$ of the words in your CoT}. This is a frozen-model, $\gamma$-aware baseline that compresses
the chain in a single forward pass without any finetuning.

\paragraph{\texttt{Full-chain}.}
The LoRA adapter is finetuned on the unmodified self-generated chains ($\gamma=1.0$, no token selection).

\paragraph{\texttt{No-chain}.}
The LoRA adapter is finetuned on the self-generated answer span only, with the entire reasoning chain stripped.

\paragraph{\texttt{Uniform}.}
For each chain, a uniformly random subset of $\lceil\gamma T\rceil$
tokens is retained at each retention budget.

\paragraph{\texttt{TokenSkip}.}
We follow \citet{xia2025tokenskip} and use a pretrained
LLMLingua-2 \citep{pan2024llmlingua} classifier as the external token scorer. The classifier assigns per-token importance values to each self-generated chain, and the top-$\lceil\gamma T\rceil$ tokens are retained at each retention budget.

\paragraph{\texttt{GoGI-$\ell_1$}.}
We use the single-layer gradient norm scorer from
Adaptive GoGI-Skip~\citep{zhuang2025accelerating}. The gradient of
$\log p_M(a \mid x, c)$ is taken at the residual state of the last
causally connected layer ($l^{\star} = \max(L-2, 0)$), and the per-token score is $\|\nabla_{h_i^{(l^{\star})}} \log p_M\|_1$. We adopt the single-layer formulation as the closest gradient-norm-only counterpart to \textsc{MIST}.

\paragraph{\texttt{Perplexity}.}
We use per-token self-information $s_i = -\log p_M(t_i \mid x, t_{<i})$
following the LLMLingua-1 prompt compression scorer
\citep{jiang2023llmlingua}; higher-self-information tokens are
retained, on the grounds that they carry more conditional information.

\paragraph{\texttt{Attention rollout}.}
Following \citet{abnar2020quantifying}, we aggregate per-head attention
weights into a cross-layer importance score. Starting from the
identity, the cumulative attention matrix at layer $l$ is
\begin{equation}
    A^{(l)} = (\widetilde{A}^{(l)} + I) \, A^{(l-1)}, \quad
    A^{(0)} = I,
\end{equation}
where $\widetilde{A}^{(l)}$ is the head-averaged attention matrix at
layer $l$. The per-token score is the column sum of $A^{(L)}$ at the
answer position(s).

\paragraph{\texttt{H2O}.}
We use the Heavy-Hitter Oracle score \citep{zhang2023ho}, defined as the cumulative attention mass received by each token across all layers and heads from the answer position(s):
\begin{equation}
    \mathrm{score}(i) =
    \sum_{l, h} \sum_{j \in \mathrm{answer}} A^{(l, h)}_{j, i}.
\end{equation}

\section{Training, Evaluation, and Implementation Details}
\label{app:protocol}
This appendix details the pipeline, hyperparameters, evaluation
protocol, and hardware configuration used in all experiments.

\paragraph{Pipeline overview.}
For each (model, dataset) cell we follow a three-stage protocol:
(i) self-generate CoT chains from the target model on the
training set and retain those whose extracted answer matches the
gold label; (ii) score each retained chain with \textsc{MIST} or
the baseline scorer, producing per-token importance values; and
(iii) for each retention budget
$\gamma \in \{0.5, 0.6, 0.7, 0.8, 0.9\}$, retain the
top-$\lceil \gamma T \rceil$ tokens of every chain, fine-tune
the target model with a LoRA adapter on the resulting compressed
chains, and evaluate by greedy decoding on the test set.

We distinguish two ratios here. The training retention budget $\gamma$ is the per-chain fraction of CoT tokens kept when constructing the SFT supervision (so $\gamma=0.7$ retains 70\% of the original chain's tokens). The reported inference compression rate is the relative reduction in total tokens generated by the fine-tuned adapter, measured against the full-chain LoRA baseline. The two need not coincide: the adapter learns its own length policy from the $\gamma$-mixture and, especially on smaller models, often produces chains shorter than the average retention floor would suggest.

\paragraph{Self-generation (Phase 0).}
We greedy-decode the target model on training-set queries using
temperature $0$ and top-$p$ $1.0$. The maximum number of new
tokens per chain is set per dataset to match each task's typical
chain length: $512$ for GSM8K, $768$ for MMLU-Pro and BBH-MC,
and $1{,}024$ for MATH. Generations are filtered to keep only
chains whose extracted answer matches the gold label.

\noindent\textit{Reproducibility and fairness.}
The same set of self-generated chains, sampled once per (model,
dataset) cell, is reused across all scorers (\textsc{MIST} and
all baselines), so any performance difference is attributable to
the scoring stage rather than to differences in supervision data.

\paragraph{Token scoring (Phase 1).}
For each retained chain, computing \textsc{MIST} requires two
backward passes through the target model: a source backward
through the answer log-likelihood
$\log p_M(a \mid x, c)$ on the full prompt-and-chain input, and
a target backward through $\log p_M(a \mid x, \emptyset)$ on the
prompt-only no-chain input. We register retain-grad hooks at the
output of every transformer block, so a single source backward
simultaneously exposes
$\nabla_{h_i^{(l)}}\log p_M^{\mathrm{src}}$ for every chain
position $i$ and every layer $l$ (Appendix~\ref{app:proofs:implementation});
the target backward analogously exposes the per-layer gradient
at the target's final position. The per-layer activations
$h_i^{(l), \mathrm{src}}$ and the logit-lens layer weight
$\bar{c}_l$ are read off the same source forward pass. 

\paragraph{LoRA fine-tuning (Phase 2).}
We follow the TokenSkip~\citep{xia2025tokenskip} fine-tuning
recipe so that any performance gap between \textsc{MIST} and a
baseline scorer can be attributed to the scoring stage alone.
LoRA adapters use rank $r = 8$, scaling $\alpha = 16$, dropout
$0.0$, and target all linear projections. Training uses learning rate
$5{\times}10^{-5}$ with linear warmup over $10\%$ of total
steps, $3$ epochs, per-device batch size $1$, gradient
accumulation $8$ (effective batch $8$), and context length
$2{,}048$. 
For each (model, dataset, scorer) triple we train a single LoRA adapter on a uniform mixture of compressed chains at $\gamma \in {0.5, 0.6, 0.7, 0.8, 0.9, 1.0}$, following TokenSkip's $\gamma$-mixing recipe \citep{xia2025tokenskip}. At inference, the same adapter is decoded under each evaluation $\gamma$ by prepending the standard \texttt{Please reduce $(1-\gamma)\times 100\%$ of the words in your CoT} directive (TokenSkip Table 1).

\paragraph{Evaluation (Phase 3).}
At evaluation, each adapter is loaded into the target model and
generation is performed by greedy decoding (temperature $0$,
top-$p$ $1.0$) on the held-out test set with evaluation batch
size $16$. The maximum number of new tokens per generation
matches the Phase 0 budget for each dataset ($512$ for GSM8K,
$768$ for MMLU-Pro and BBH-MC, $1{,}024$ for MATH). 

\paragraph{Hardware and software.}
All experiments run on $2 \times$ NVIDIA A100 (80~GB) GPUs.
The implementation uses PyTorch 2.9, Hugging Face Transformers,
and the PEFT library for LoRA fine-tuning.
The pipeline incurs the
following one-time cost: self-generation of reasoning chains
takes 0.5--2 GPU-hours; scoring chains with \textsc{MIST}
requires two backward passes per chain and takes 0.3--1.0
GPU-hours; LoRA adapter training takes 1--3 GPU-hours; and
evaluation at five retention budgets takes 0.5--1.5 GPU-hours.
Aggregated across all four models, four datasets, and ten
methods (six scorers plus four trivial baselines), the total
compute reported in this paper is on the order of 1{,}500
GPU-hours.

\section{Taylor Expansions and Remainder Bounds}
\label{app:proofs}
This appendix formalizes the first-order Taylor expansions of the per-layer necessity and sufficiency terms $\widehat\phi_i^{(l)}$ and $\widehat\psi_i^{(l)}$ and
gives explicit per-layer remainder bounds, completing the technical content deferred from Section~\ref{sec:method:taylor}

\subsection{Setup and notation}
\label{app:proofs:setup}

\paragraph{Local per-layer functions.}
Let $M$ be a transformer language model with $L$ residual-stream
layers and hidden dimension $d$. For each chain position $i \in [T]$
and layer $l \in [L]$, define the source-side local function
\begin{equation}
\resizebox{0.98\columnwidth}{!}{$
    f^{\mathrm{src}}_{i,l}(\mathbf{u})
    :=
    \log p_M\bigl(a \mid x, c;\; h_i^{(l)} \leftarrow \mathbf{u}\bigr),
    \quad \mathbf{u} \in \mathbb{R}^d
$}
\end{equation}
where the semicolon notation indicates that the residual-stream state
$h_i^{(l)}$ is replaced by $\mathbf{u}$, after which the remaining
computation is evaluated normally. At the natural value the
intervened computation coincides with the standard forward pass, and
its derivative with respect to $\mathbf{u}$ coincides with the
reverse-mode gradient of $\log p_M^{\mathrm{src}}$ with respect to
$h_i^{(l)}$:
\begin{equation}
    \nabla f^{\mathrm{src}}_{i,l}\bigl(h_i^{(l), \mathrm{src}}\bigr)
    \;=\;
    \nabla_{h_i^{(l)}} \log p_M^{\mathrm{src}}.
    \label{eq:autograd-equality-src}
\end{equation}
The target-side local function is defined analogously:
\begin{equation}
\resizebox{0.98\columnwidth}{!}{$
    f^{\mathrm{tgt}}_{l}(\mathbf{u}) \;:=\;
    \log p_M\bigl(a \mid x, \emptyset;\;
        h_{\mathrm{final}}^{(l)} \leftarrow \mathbf{u}\bigr),
    \quad \mathbf{u} \in \mathbb{R}^d,
    $}
\end{equation}
with the analogous gradient identity
\begin{equation}
    \nabla f^{\mathrm{tgt}}_{l}\bigl(h_{\mathrm{final}}^{(l), \mathrm{tgt}}\bigr)
    \;=\;
    \nabla_{h_{\mathrm{final}}^{(l)}} \log p_M^{\mathrm{tgt}}.
    \label{eq:autograd-equality-tgt}
\end{equation}

For brevity, write
\(\log p_M^{\mathrm{src}} := \log p_M(a\mid x,c)\) and
\(\log p_M^{\mathrm{tgt}} := \log p_M(a\mid x,\emptyset)\).

\paragraph{Regularity assumption.}
We assume each $f^{\mathrm{src}}_{i,l}$ and $f^{\mathrm{tgt}}_{l}$ is locally $C^{2}$ on a neighborhood containing the unperturbed point and the corresponding perturbed point. This is a working assumption invoked only to state Lagrange-form Taylor remainders; the empirical method requires only first-order gradients.

\subsection{Necessity: Taylor expansion of $\phi_i^{(l)}$}
\label{app:proofs:phi}
The per-layer necessity intervention at layer $l$ replaces
$h_i^{(l)}$ with $\mathbf{0}$ and lets subsequent layers evaluate
normally from this perturbation. The corresponding per-layer saliency
and displacement are
\begin{equation}
\begin{aligned}
    \phi_i^{(l)}
    &:=
    f^{\mathrm{src}}_{i,l}\bigl(h_i^{(l), \mathrm{src}}\bigr)
    -
    f^{\mathrm{src}}_{i,l}\bigl(\mathbf{0}\bigr),
    \\
    \delta^{\phi}_{i,l}
    &:=
    -\,h_i^{(l), \mathrm{src}}
    \in \mathbb{R}^d .
\end{aligned}
\end{equation}

\paragraph{Expansion.}
By Taylor's theorem with Lagrange remainder applied to
$f^{\mathrm{src}}_{i,l}$ along the segment from
$h_i^{(l), \mathrm{src}}$ to $\mathbf{0}$,
\begin{equation}
\begin{aligned}
    f^{\mathrm{src}}_{i,l}(\mathbf{0})
    &=
    f^{\mathrm{src}}_{i,l}\bigl(h_i^{(l), \mathrm{src}}\bigr)
    +
    \bigl\langle
        \nabla f^{\mathrm{src}}_{i,l}
        \bigl(h_i^{(l), \mathrm{src}}\bigr),\;
        \delta^{\phi}_{i,l}
    \bigr\rangle
    \\
    &\quad
    +
    \tfrac{1}{2}
    \bigl(\delta^{\phi}_{i,l}\bigr)^{\!\top}
    \nabla^2 f^{\mathrm{src}}_{i,l}
    \bigl(\xi^{\phi}_{i,l}\bigr)
    \delta^{\phi}_{i,l}.
\end{aligned}
\end{equation}
for some $\xi^{\phi}_{i,l}$ on the segment. Substituting
$\delta^{\phi}_{i,l} = -h_i^{(l), \mathrm{src}}$, using
Eq.~\eqref{eq:autograd-equality-src}, and rearranging yields
\begin{equation}
        \phi_i^{(l)} \;=\;
        \widehat\phi_i^{(l)} \;-\; R^{\phi}_{i,l}
    \label{eq:phi-per-layer-identity}
\end{equation}
where
\begin{align}
    \widehat\phi_i^{(l)}
    &\;=\;
    \bigl\langle
        \nabla_{h_i^{(l)}} \log p_M^{\mathrm{src}},\;
        h_i^{(l), \mathrm{src}}
    \bigr\rangle,
    \\
    R^{\phi}_{i,l}
    &\;=\;
    \tfrac{1}{2}\,
    \bigl(\delta^{\phi}_{i,l}\bigr)^{\!\top}
    \nabla^2 f^{\mathrm{src}}_{i,l}\bigl(\xi^{\phi}_{i,l}\bigr)
    \delta^{\phi}_{i,l}.
\end{align}
Eq.~\eqref{eq:phi-per-layer-identity} recovers the per-layer
first-order term defined in Eq.~(\ref{eq:phi-layer}) of the main
text.

\paragraph{Remainder bound.}
Let
\begin{equation}
    \beta^{\phi}_{i,l} := \sup_{\xi \in \mathcal{S}^{\phi}_{i,l}}
\|\nabla^2 f^{\mathrm{src}}_{i,l}(\xi)\|_{\mathrm{op}}
\end{equation}
denote the Hessian operator-norm bound on the segment
$\mathcal{S}^{\phi}_{i,l}$ from $h_i^{(l), \mathrm{src}}$ to
$\mathbf{0}$. Then
\begin{equation}
    \bigl|R^{\phi}_{i,l}\bigr| \;\leq\;
    \tfrac{1}{2}\, \beta^{\phi}_{i,l}\,
    \bigl\|h_i^{(l), \mathrm{src}}\bigr\|_2^{2}.
    \label{eq:phi-remainder-bound}
\end{equation}

\subsection{Sufficiency: Taylor expansion of $\psi_i^{(l)}$}
\label{app:proofs:psi}

The per-layer sufficiency intervention at layer $l$ replaces the
target's final-position residual at layer $l$ with the source's
position-$i$ residual at the same layer, and lets subsequent layers
evaluate normally from this perturbation. The corresponding per-layer
saliency and displacement are
\begin{equation}
\begin{aligned}
    \psi_i^{(l)}
    &:=
    f^{\mathrm{tgt}}_{l}\bigl(h_i^{(l), \mathrm{src}}\bigr)
    -
    f^{\mathrm{tgt}}_{l}\bigl(h_{\mathrm{final}}^{(l), \mathrm{tgt}}\bigr),
    \\
    \delta^{\psi}_{i,l}
    &:=
    h_i^{(l), \mathrm{src}}
    -
    h_{\mathrm{final}}^{(l), \mathrm{tgt}}
    \in \mathbb{R}^d .
\end{aligned}
\end{equation}

\paragraph{Expansion.}
By Taylor's theorem with Lagrange remainder applied to
$f^{\mathrm{tgt}}_{l}$ along the segment from
$h_{\mathrm{final}}^{(l), \mathrm{tgt}}$ to $h_i^{(l), \mathrm{src}}$,
\begin{equation}
\resizebox{0.98\columnwidth}{!}{$
\begin{aligned}
    f^{\mathrm{tgt}}_{l}\bigl(h_i^{(l), \mathrm{src}}\bigr)
    &=
    f^{\mathrm{tgt}}_{l}\bigl(h_{\mathrm{final}}^{(l), \mathrm{tgt}}\bigr)
    +
    \bigl\langle
        \nabla f^{\mathrm{tgt}}_{l}
        \bigl(h_{\mathrm{final}}^{(l), \mathrm{tgt}}\bigr),\;
        \delta^{\psi}_{i,l}
    \bigr\rangle
    \\
    &\quad
    +
    \tfrac{1}{2}
    \bigl(\delta^{\psi}_{i,l}\bigr)^{\!\top}
    \nabla^2 f^{\mathrm{tgt}}_{l}
    \bigl(\xi^{\psi}_{i,l}\bigr)
    \delta^{\psi}_{i,l}.
\end{aligned}
$}
\end{equation}
for some $\xi^{\psi}_{i,l}$ on the segment. Using
Eq.~\eqref{eq:autograd-equality-tgt} and rearranging yields
\begin{equation}
        \psi_i^{(l)} \;=\;
        \widehat\psi_i^{(l)} \;+\; R^{\psi}_{i,l}
    \label{eq:psi-per-layer-identity}
\end{equation}
where
\begin{align}
    \widehat\psi_i^{(l)}
    &\;=\;
    \bigl\langle
        \nabla_{h_{\mathrm{final}}^{(l)}} \log p_M^{\mathrm{tgt}},\;
        h_i^{(l), \mathrm{src}} - h_{\mathrm{final}}^{(l), \mathrm{tgt}}
    \bigr\rangle,
    \\
    R^{\psi}_{i,l}
    &\;=\;
    \tfrac{1}{2}\,
    \bigl(\delta^{\psi}_{i,l}\bigr)^{\!\top}
    \nabla^2 f^{\mathrm{tgt}}_{l}\bigl(\xi^{\psi}_{i,l}\bigr)
    \delta^{\psi}_{i,l}.
\end{align}
Eq.~\eqref{eq:psi-per-layer-identity} recovers the per-layer
first-order term defined in Eq.~(\ref{eq:psi-layer}) of the main
text. The plus sign on $R^{\psi}_{i,l}$ (as opposed to the minus
sign on $R^{\phi}_{i,l}$) appears because sufficiency is defined as
a likelihood gain whereas necessity is defined as a likelihood drop.

\paragraph{Remainder bound.}
Let
\begin{equation}
    \beta^{\psi}_{i,l} := \sup_{\xi \in \mathcal{S}^{\psi}_{i,l}}
\|\nabla^2 f^{\mathrm{tgt}}_{l}(\xi)\|_{\mathrm{op}}
\end{equation}
denote the Hessian operator-norm bound on the segment
$\mathcal{S}^{\psi}_{i,l}$. Then
\begin{equation}
    \bigl|R^{\psi}_{i,l}\bigr| \;\leq\;
    \tfrac{1}{2}\, \beta^{\psi}_{i,l}\,
    \bigl\|h_i^{(l), \mathrm{src}}
        - h_{\mathrm{final}}^{(l), \mathrm{tgt}}\bigr\|_2^{2}.
    \label{eq:psi-remainder-bound}
\end{equation}

\subsection{Recovering all per-layer gradients from a single backward
    pass}
\label{app:proofs:implementation}

Computing $\widehat\phi_i^{(l)}$ for every $(i, l)$ pair naively
suggests $TL$ separate backward passes. In practice, a single
backward through $\log p_M^{\mathrm{src}}$ simultaneously recovers
all gradients $\nabla_{h_i^{(l)}} \log p_M^{\mathrm{src}}$ by
registering retain-grad hooks on each transformer block's output
residual; reverse-mode autograd propagates gradients to all
intermediate activations through the standard chain rule. The hooks
expose these gradients without modifying the computation graph. Thus
all necessity terms require one source forward and one backward pass
through $M$, and all sufficiency terms require one target forward
and one backward pass, each up to activation-caching overhead.

\subsection{Empirical Tightness of the First-Order Term}
\begin{figure}[h]
\centering
  \includegraphics[width=0.7\columnwidth]{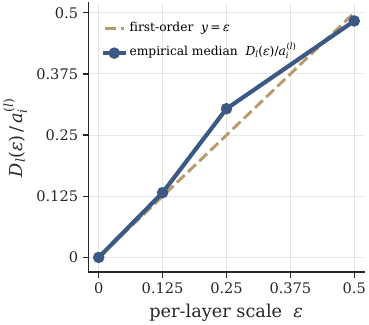}
  \caption{Empirical tightness of the per-layer first-order Taylor term.}
  \label{fig:m1-tightness}
\end{figure}

The analytic bounds in \S\ref{app:proofs:phi}--\ref{app:proofs:psi} are worst-case statements parametrised by an unspecified Hessian constant; we close that gap by measuring directly how tightly the first-order term predicts the true ablation drop on real chains.
For one chain token $i$ at one residual block $l$, we scale the residual stream $h_i^{(l)} \!\to\! (1-\varepsilon)\,h_i^{(l)}$ at that block only (others unperturbed) and measure the answer-log-likelihood drop
\begin{equation}
\resizebox{0.98\columnwidth}{!}{$
    D_l(\varepsilon) := \log p_M(a\!\mid\!x,c)
    - \log p_M\!\bigl(a\!\mid\!x, c;\,
        h_i^{(l)} \!\leftarrow\! (1-\varepsilon)\,h_i^{(l)}\bigr).
$}
\end{equation}
The per-layer Taylor main term (Eq.~\ref{eq:phi-layer}),
$a_i^{(l)} := \langle \nabla_{h_i^{(l)}}\!\log p_M^{\mathrm{src}},\,
h_i^{(l)} \rangle = \widehat\phi_i^{(l)}$,
yields the first-order prediction
$D_l(\varepsilon) \approx \varepsilon a_i^{(l)}$, so the normalized drop
$D_l(\varepsilon)/a_i^{(l)}$ equals $\varepsilon$ in the first-order-tight limit.
Measured on $8$ GSM8K chains with Qwen2.5-1.5B-Instruct in fp32 over
functionally active $(i,l)$ cells (top quartile of $|D_l(\varepsilon = 1)|$
per chain, floored at $10^{-5}$ nats), the empirical median tracks
$y=\varepsilon$ to within $\lesssim 10\%$ overall (Figure~\ref{fig:m1-tightness}).

\subsection{Empirical Ranking Fidelity of the Taylor Proxies Against the Exact Interventions}
\label{app:proofs:ranking-fidelity}

We further evaluate selection fidelity by comparing the top-ranked tokens under each Taylor proxy with those selected by its exact intervention on GSM8K using Qwen2.5-1.5B-Instruct. As shown in Table~\ref{tab:ranking-fidelity}, necessity achieves top-$\gamma$ agreement of $0.72$ at $\gamma=0.3$ and $0.81$ at $\gamma=0.5$, while sufficiency reaches $0.68$ and $0.76$, respectively. All values substantially exceed the corresponding chance levels of $0.3$ and $0.5$, indicating that both proxies reliably preserve the token selections induced by their exact interventions.

\begin{table}[h]
\centering
\small
\begin{tabular}{@{}lcc@{}}
  \toprule
  Axis & top@0.3 & top@0.5 \\
  \midrule
  Necessity $\widehat\phi$ & $0.72$ & $0.81$ \\
  Sufficiency $\widehat\psi$ & $0.68$ & $0.76$ \\
  \bottomrule
\end{tabular}
\caption{Top-$\gamma$ agreement between the first-order Taylor proxies ($\widehat\phi,\widehat\psi$) and the exact interventions ($\phi,\psi$) on GSM8K using Qwen2.5-1.5B-Instruct. Chance agreement is $\gamma$.}
\label{tab:ranking-fidelity}
\end{table}

\section{Additional Main Results on BBH and MMLU-Pro}
\label{app:additional-results}

Table~\ref{tab:bbh-mmlu-additional-results} reports the per-(model, scorer) compression numbers underlying Figs.~\ref{fig:bbh} and~\ref{fig:mmlu}, in the same $\Delta$Acc / compression-rate format as Table~\ref{tab:main-compression-results}.

\begin{table}[t]
\centering
\renewcommand{\arraystretch}{1.36}
\resizebox{\columnwidth}{!}{%
\begin{tabular}{c c l c c}
\toprule
\textbf{Dataset} & \textbf{Model} & \textbf{Method}
& \textbf{$\Delta$ Acc. (\%)} & \textbf{Comp. rate} \\
\midrule

\multirow{5}{*}{BBH}
& \multirow{5}{*}{\rotatebox{90}{Llama-3.1-8B-Inst.}}
& $\star$ MIST  & $\uparrow 3.7$    & $11.6\%$  \\
& & tokenskip    & $\uparrow 1.5$    & $4.6\%$  \\
& & gogi\_l1     & $\downarrow 1.2$  & $6.8\%$ \\
& & attn\_rollout& $\downarrow 0.7$  & $6.1\%$  \\
& & h2o          & $\uparrow 0.2$  & $7.8\%$  \\

\midrule

\multirow{10}{*}{MMLU-Pro}
& \multirow{5}{*}{\rotatebox{90}{Qwen2.5-1.5B-Inst.}}
& $\star$ MIST  & $\downarrow 2.8$  & $20.8\%$ \\
& & tokenskip    & $\downarrow 4.3$  & $-0.5\%$ \\
& & gogi\_l1     & $\downarrow 6.6$  & $3.4\%$  \\
& & perplexity   & $\downarrow 7.1$  & $7.8\%$  \\
& & h2o          & $\downarrow 3.5$  & $12.7\%$ \\

\cmidrule(lr){2-5}

& \multirow{5}{*}{\rotatebox{90}{Llama-3.1-8B-Inst.}}
& $\star$ MIST  & $\downarrow 1.0$  & $3.4\%$  \\
& & tokenskip    & $\downarrow 2.5$  & $1.2\%$  \\
& & gogi\_l1     & $\downarrow 2.6$  & $2.1\%$  \\
& & perplexity   & $\downarrow 3.7$  & $6.9\%$  \\
& & h2o          & $\downarrow 3.4$  & $2.5\%$  \\

\bottomrule
\end{tabular}%
}
\caption{
Additional main results on BBH and MMLU-Pro, in the same format as Table~\ref{tab:main-compression-results}. Each entry is the mean over $\gamma \in \{0.6, 0.7, 0.8, 0.9\}$ relative to the Full CoT baseline. $\Delta$Acc.\,(\%): relative accuracy change ($\downarrow$ drop, $\uparrow$ gain); Comp.\,rate: fraction by which the LoRA-adapted model's inference-time generated tokens are reduced relative to the same Full CoT baseline. A negative compression rate indicates that the compressed-supervision adapter generated \emph{more} tokens than the Full CoT baseline at evaluation.
}
\label{tab:bbh-mmlu-additional-results}
\end{table}

\section{Direct inference-cost measurement.}
\label{app:inference-cost}

\begin{table}[t]
  \centering
  \small
  \setlength{\tabcolsep}{6pt}
  \begin{tabular}{lccc}
    \toprule
    & \multicolumn{2}{c}{Latency (s)} & \\
    \cmidrule(lr){2-3}
    Model & $\gamma{=}1.0$ & $\gamma{=}0.5$ & Speedup \\
    \midrule
    Qwen2.5-1.5B & $1.44$ & $1.18$ & $1.22\times$ \\
    Llama-3.1-8B & $1.38$ & $1.06$ & $1.30\times$ \\
    Mistral-7B   & $1.72$ & $1.20$ & $1.43\times$ \\
    \bottomrule
  \end{tabular}
  \caption{Directly measured inference cost of the \textsc{MIST}
    adapter on GSM8K at the most aggressive trained budget
    ($\gamma{=}0.5$) versus the uncompressed reference
    ($\gamma{=}1.0$, the same adapter with no compression
    directive). Latency is the mean wall-clock decode time per
    problem at batch size $16$; speedup is the latency ratio
    $\gamma{=}1.0$ over $\gamma{=}0.5$.}
  \label{tab:inference-cost}
\end{table}

The main results use compression rate as a hardware-agnostic proxy for efficiency. To quantify realized serving gains, we additionally report wall-clock decoding latency on the full GSM8K test set ($1{,}319$ examples), using batched greedy decoding with batch size $16$, \texttt{bfloat16}, and SDPA. For each model, we compare the \textsc{MIST} adapter at the most aggressive trained budget ($\gamma{=}0.5$) with the same adapter under the no-compression directive ($\gamma{=}1.0$), which serves as the Full-CoT reference. As shown in Table~\ref{tab:inference-cost}, compression reduces latency from $1.44$ to $1.18$ seconds for Qwen2.5-1.5B-Instruct, from $1.38$ to $1.06$ seconds for Llama-3.1-8B-Instruct, and from $1.72$ to $1.20$ seconds for Mistral-7B-Instruct-v0.3, corresponding to speedups of $1.22\times$, $1.30\times$, and $1.43\times$, respectively. These results confirm that shorter reasoning traces translate into measurable end-to-end inference savings across models.

\section{Per-Axis Normalization Before Combining}
\label{app:proofs:normalize}
The necessity and sufficiency scores $\widehat\phi_i$ and
$\widehat\psi_i$ (Eq.~\ref{eq:axis-agg}) may differ in scale.
To balance their contributions, we apply standard z-score
normalization~\cite{jain2005score} to each axis independently, using the mean
and standard deviation computed over tokens within each chain.
The unified \textsc{MIST} score in Eq.~\eqref{eq:mist-score}
is then computed using the normalized scores.

\section{Empirical Per-Layer Logit-Lens Weight}
\label{app:proofs:layerweight}
\begin{figure}[h]
\centering
  \includegraphics[width=\columnwidth]{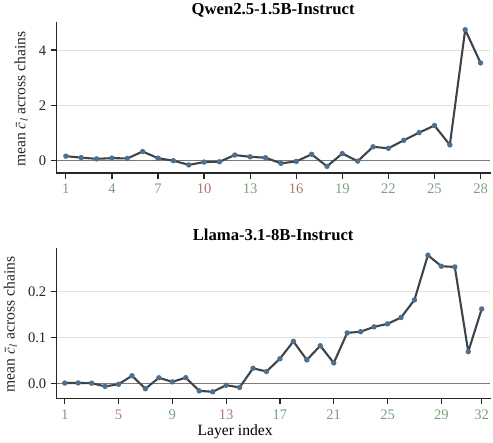}
  \caption{Empirical mean of the logit-lens layer weight $\bar c_l$ (Eq.~\ref{eq:logit-lens-weight}) across $30$ BBH chains for two of the paper-headline target models. Weight magnitude is overwhelmingly concentrated in the late layers, while many middle layers carry near-zero weight; both observations are direct empirical support for the layer-weighted aggregation in Eq.~\eqref{eq:axis-agg} over a uniform per-layer weight.}
  \label{fig:bar-c-l-empirical}
\end{figure}
The layer weight $\bar c_l$ in Eq.~\eqref{eq:logit-lens-weight} is the inner product of layer~$l$'s residual-stream update with the answer unembedding direction, chain-averaged over the chain tokens. Figure~\ref{fig:bar-c-l-empirical} plots its mean across $30$ BBH chains layer by layer for Qwen2.5-1.5B-Instruct and Llama-3.1-8B-Instruct, two of the four main target models.

The figure makes the following empirical points concrete. First, $\bar c_l$ is far from uniform across depth.
This explains why the \textsc{MIST}-uniform ablation in \S\ref{sec:exp:ablation} (Fig.~\ref{fig:component_ablation}) loses up to $4.6$~pp --- uniform weighting wastes selection budget on layer-token cells whose updates are essentially answer-neutral.
Besides, the late-layer dominance is consistent across the two model families, so the aggregation in Eq.~\eqref{eq:axis-agg} is not specific to a single architecture.

\section{Licenses and Terms of Use}
\label{app:licenses}

This section discusses the licenses and terms of use for all
external artifacts used in our experiments and for the artifacts
we release. All datasets and models we use are publicly available
under permissive research-friendly licenses, and our use is
consistent with their intended research purposes.

\paragraph{Datasets.}
We use four publicly released reasoning benchmarks. GSM8K
\citep{cobbe2021training} is released under the MIT License. The
MATH dataset \citep{hendrycks2021measuring} is released under the
MIT License. MMLU-Pro \citep{wang2024mmlupro} is released under
the MIT License via the TIGER-Lab repository on Hugging Face.
BIG-Bench Hard \citep{suzgun2023challenging} is released under
the Apache License~2.0 as part of the BIG-Bench suite. All four
datasets are distributed for research use and our experimental
use is consistent with their intended scope. We do not
redistribute the underlying data; experiments load the datasets
from their original sources.

\paragraph{Models.}
We use four publicly released instruction-tuned models.
Qwen2.5-1.5B-Instruct and Qwen2.5-7B-Instruct \citep{qwen2.5}
are released under the Qwen License (Apache~2.0--based).
Llama-3.1-8B-Instruct \citep{grattafiori2024llama} is released
under the Llama~3.1 Community License Agreement. Mistral-7B-
Instruct-v0.3 \citep{Jiang2023Mistral7} is released under the
Apache License~2.0. All four models permit research use,
fine-tuning, and the release of derived artifacts (such as LoRA
adapters) under their respective terms.

\paragraph{Consistency with intended use.}
Our use of all external artifacts is consistent with their
intended research purposes. The four reasoning datasets (GSM8K,
MATH, MMLU-Pro, BIG-Bench Hard) are released as research
benchmarks for evaluating multi-step reasoning in language
models, and our use is restricted to benchmarking and training
LoRA adapters on the released training splits. The four
instruction-tuned models (Qwen2.5-1.5B/7B, Llama-3.1-8B,
Mistral-7B-Instruct-v0.3) are released to support research,
fine-tuning, and the development of derived artifacts under
their respective terms. Adapters derived from each base model
inherit the original access conditions of that base model;
in particular, adapters derived from Llama-3.1-8B-Instruct are
released only for use with the Llama-3.1 base model, and are not
used to train, distill into, or otherwise improve language
models outside the Llama family, consistent with the Llama~3.1
Community License. All other adapters (built on Qwen2.5 and
Mistral, both released under Apache~2.0) carry no such derivative
restrictions. All experimental use is for research purposes
only; we do not deploy the trained adapters in commercial
products.

\end{document}